\documentclass[10pt,twocolumn,letterpaper]{article}

\usepackage[pagenumbers]{iccv} 

\usepackage{comment}

\usepackage{multirow}
\usepackage{multicol}
\usepackage{array}
\newcolumntype{C}[1]{>{\centering\arraybackslash}p{#1\dimexpr1.5cm}}

\usepackage{framed}
\usepackage{wrapfig}

\definecolor{iccvblue}{rgb}{0.21,0.49,0.74}
\usepackage[pagebackref,breaklinks,colorlinks,allcolors=iccvblue]{hyperref}

\def\paperID{****} 
\def\confName{ICCV}
\def\confYear{2025}

\title{Think Before You Segment:\\ High-Quality Reasoning Segmentation with GPT Chain of Thoughts}

\author{Shiu-hong Kao\\
HKUST\\
{\tt\small skao@cse.ust.hk}
\and
Yu-Wing Tai\\
Dartmouth College\\
{\tt\small yu-wing.tai@dartmouth.edu}
\and
Chi-Keung Tang\\
HKUST\\
{\tt\small cktang@cs.ust.hk}
}

\begin{document}
\twocolumn[{%
\renewcommand\twocolumn[1][]{#1}%
\maketitle
\centering
\includegraphics[width=\linewidth]{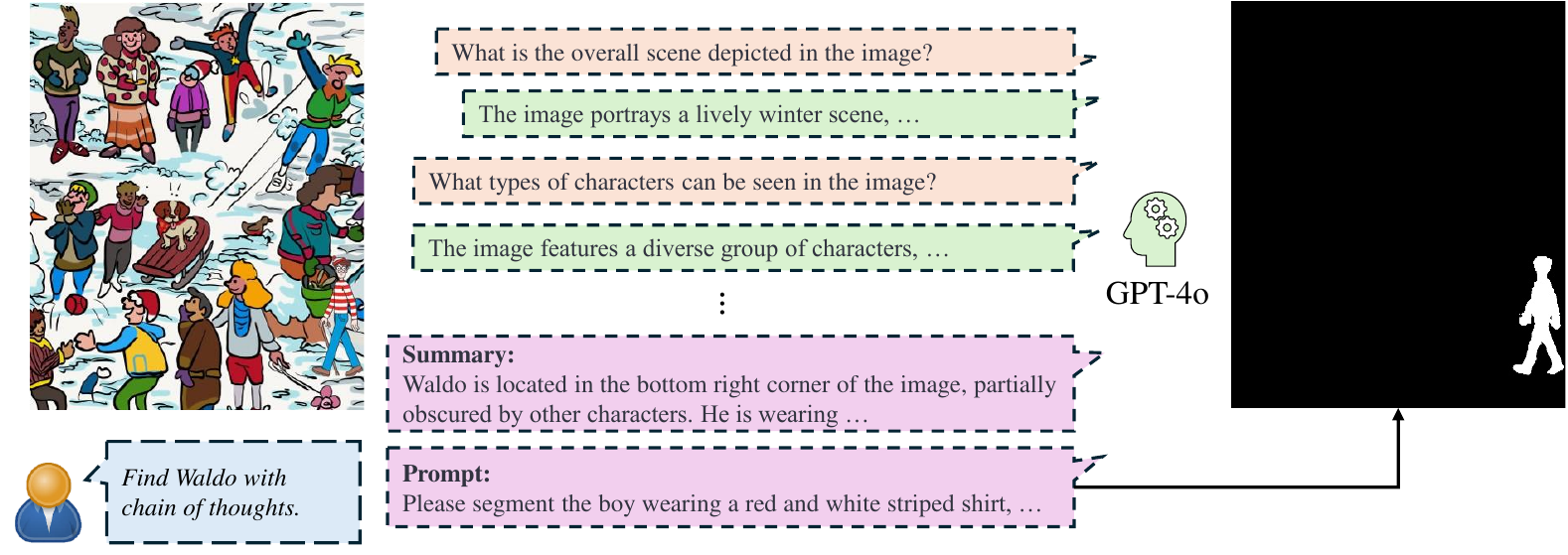}
\vspace{-0.2in}
\captionof{figure}{We propose \textbf{ThinkFirst}, a novel Chain-of-Thought (CoT) reasoning segmentation framework that generates an accurate object mask given a text prompt, implicit or explict with complex details alike, after autonomously Thinking First with GPT-4o's CoT. Our zero-shot-CoT framework can  
handle difficult scenarios such as implicit queries, camouflaged objects, out-of-domain objects with easy control. \vspace{1em}}
\label{fig:teaser}
}]
\begin{abstract}
Reasoning segmentation is a challenging vision-language task that aims to output the segmentation mask with respect to a complex,  implicit, and even non-visual query text. Previous works incorporated multimodal Large Language Models (MLLMs) with segmentation models to approach the difficult problem. However, their segmentation quality often falls short in complex cases, particularly when dealing with out-of-domain objects with intricate structures, blurry boundaries, occlusions, or high similarity with surroundings. In this paper, we introduce \textbf{ThinkFirst}, a training-free reasoning segmentation framework that leverages GPT's chain of thought to address these challenging cases. Our  approach allows GPT-4o or other powerful MLLMs to generate a detailed, chain-of-thought description of an image. This summarized description is then passed to a language-instructed segmentation assistant to aid the segmentation process. Our framework allows users to easily interact with the segmentation agent using multimodal inputs, such as easy text and image scribbles, for successive refinement or communication. We evaluate the performance of ThinkFirst on diverse objects. Extensive experiments show that, this zero-shot-CoT approach significantly improves the vanilla reasoning segmentation agent, both qualitatively and quantitatively, while being less sensitive or critical to user-supplied prompts after Thinking First. Code will be released at \href{https://github.com/DanielSHKao/ThinkFirst}{github.com/DanielSHKao/ThinkFirst}.

\end{abstract}    
\begin{figure*}
    \centering
    \includegraphics[width=\linewidth]{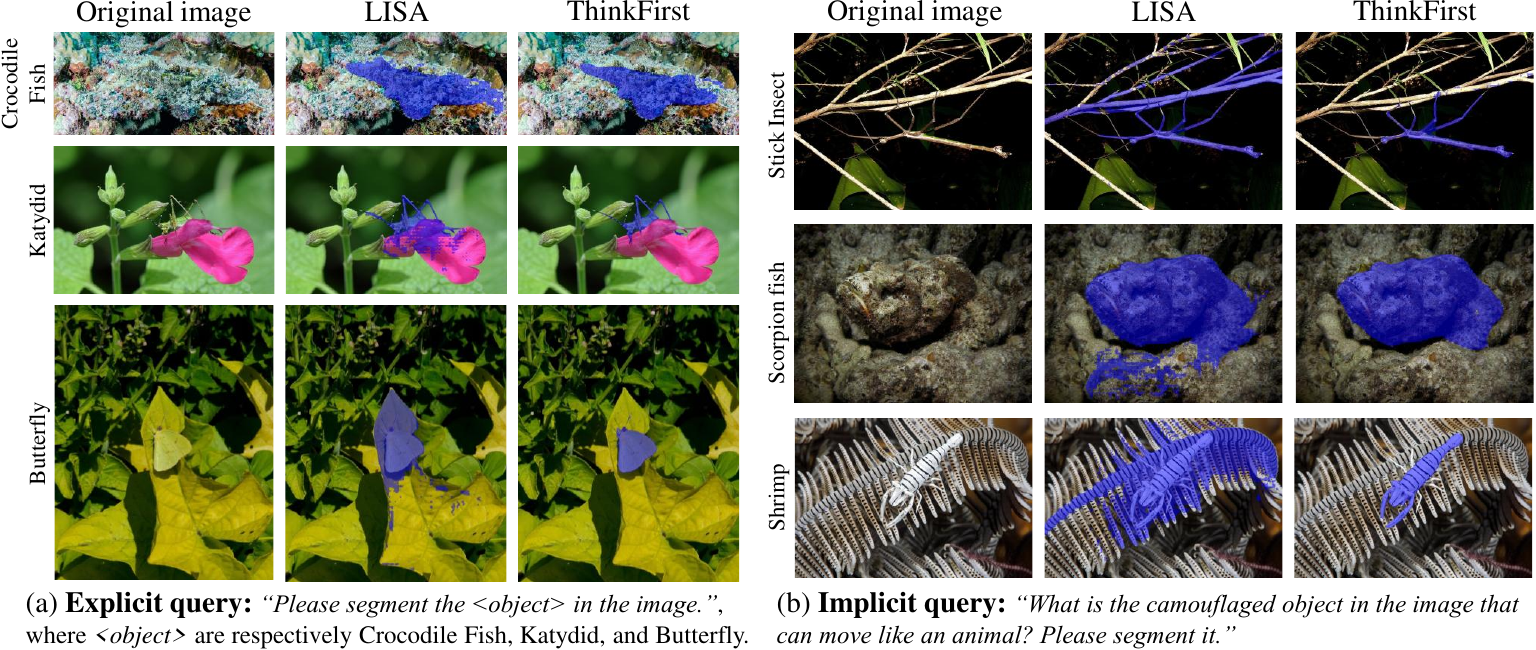}
    \vspace{-0.25in}
    \caption{Qualitative comparison. View in colors and zoom in for details. Refer to the appendix for more challenging examples.
    }
    \vspace{-0.1in}
    \label{fig:qual}
\end{figure*}

\vspace{-0.2in}
\section{Introduction}
Reasoning segmentation 
is an emerging vision-language approach 
that requires the pertinent model to generate precise image segmentation masks of objects of interest,  given an implicit text queries like \textit{``What is the largest object in the image? Please segment it.''} Unlike traditional vision-language tasks that directly associate visual data with textual descriptions, reasoning segmentation requires advanced cognitive capabilities. This is because the implicit nature of the queries often demands models to interpret nuanced instructions and engage in higher-order reasoning to accurately delineate relevant regions within the image.

\begin{figure}
    \centering
    \includegraphics[width=0.99\linewidth]{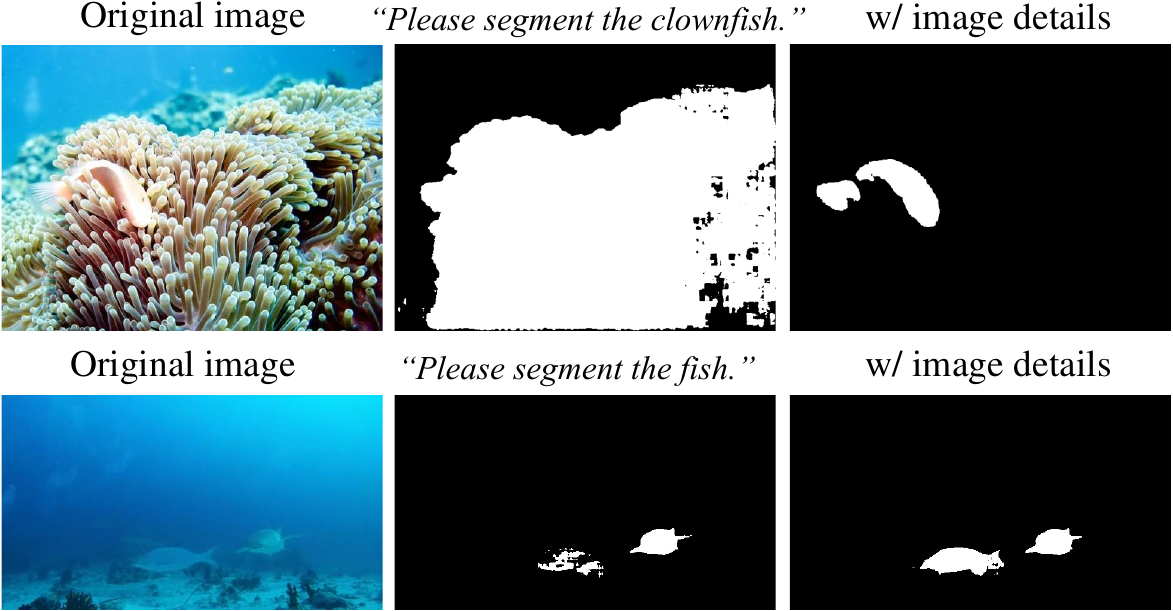}
    \vspace{-0.1in}
    \caption{ThinkFirst produces accurate results (right), while LISA fails to generate reasonable masks (middle) when given little information, especially for camouflaged objects or underwater images with severe color shift and blurry object boundaries.}
    \vspace{-0.15in}
    \label{fig:motivation}
\end{figure}
In this paper, we propose \textbf{ThinkFirst}, a simple yet effective framework that utilizes chain-of-thought (CoT) to engage in such higher-level reasoning, leading to higher-quality segmentation.
Instead of directly feeding the user query and the image to a reasoning segmentation model, 
ThinkFirst  {\em first} 
thoroughly analyzes the image with {\em chain of thoughts} using carefully designed prompts. Then, based on the CoT responses, 
a detailed summary of the image is generated,  which 
is combined with the user query to guide the reasoning segmentation model (See Figure~\ref{fig:teaser}).
To our knowledge, ThinkFirst is one of the first to use CoTs for general, zero-shot reasoning segmentation, given text and/or image scribbles queries. This zero-shot-CoT approach is training-free and is compatible with any reasoning segmentation modules, which may be applied as a plug-in to further improve their results.

\begin{figure}
    \centering
    {
    \includegraphics[width=\linewidth]{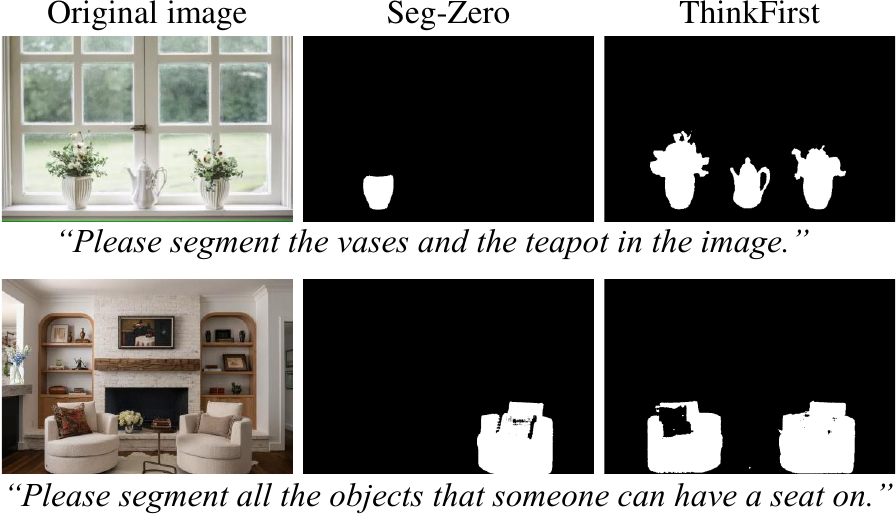}
    \vspace{-0.25in}
    }
    \caption{Comparison with Seg-Zero~\cite{liu2025seg}. Seg-Zero requires a  bounding box prior for the mask output, and thereby is limited to single target object. Instead, our zero-shot ThinkFirst can output high-quality mask for multiple objects in one pass.}
    \vspace{-0.15in}
    \label{fig:segzero}
\end{figure}
Recently, Multimodal Large Language Models (MLLMs)~\cite{alayrac2022flamingo,li2023blip,ye2023mplug,zhang2023llama} have been widely used to solve reasoning segmentation due to their excellence in image understanding~\cite{liu2023visual}, complex reasoning~\cite{chung2024scaling} and instruction following~\cite{ouyang2022training}. These models have shown remarkable capabilities in interpreting and processing visual data alongside textual inputs, making them ideal for tasks that require both visual and linguistic understanding. To generate an image-based output, previous works often directly connect an MLLM with the auto-decoder architecture of pre-trained visual backbone, e.g., SAM~\cite{kirillov2023segment} for  reasoning segmentation. The  representative work LISA~\cite{lai2024lisa} 
expands the vocabulary of MLLM with a segmentation-specific token 
and directly sends its latent features 
to a mask decoder to guide the decoder with cross-attention mechanism, \textcolor{black}{which can be regarded as a strong MLLM + SAM baseline without CoT reasoning.} 
Despite impressive results to date,
they mainly focus on images with a primary foreground or objects with clear boundaries, while still struggling to output reasonable masks for complex images with multiple objects, especially when the target has similar colors to its surroundings, 
with ambiguous boundaries or boundaries blending into the background, see Figures~\ref{fig:qual} to~\ref{fig:overview}, indicating a higher level of visual understanding is necessary for resolving ambiguities and delineating subtle yet important features.

To improve the accuracy of reasoning segmentation, a concurrent work Seg-Zero~\cite{liu2025seg} leverages a reinforcement learning approach to incorporate CoT reasoning into the multimodal LLM model. However, due to the requirement of a bounding box prior, Seg-Zero fails to generate a mask for multiple target objects (Refer to Figure~\ref{fig:segzero}).  Additionally, the notorious time cost for RL-based approaches makes the Seg-Zero less favorable for downstream applications. In this context, we claim that a training-free framework for high-quality and zero-shot reasoning segmentation is still needed to address the limitations of existing methods.

To validate our CoT approach  showcasing ThinkFirst is a clear winner in challenging cases
not adequately covered by conventional datasets,  
we qualitatively and quantitatively compare the segmentation quality in difficult camouflaged images using COD-10K~\cite{fan2020camouflaged} and underwater images using Marine Video Kit~\cite{MVK} without fine-tuning. 
We also evaluate the generalization ability of ThinkFirst on diverse out-of-domain data, from modern architecture, mid-century furniture with bold design, to application on ``Where's Waldo'' images. 
Results show that ThinkFirst indeed significantly improves the reasoning segmentation quality in difficult scenarios and given highly implicit user prompts.

\section{Related Work}
\subsection{Multimodal Large Language Model} Inspired by the remarkable reasoning ability of Large Language Models (LLM), Multimodal LLMs (MLLM) aim to incorporate visual modules for multimodal perception and reasoning, typically used to generate captions or answer questions for visual inputs. Training-free approaches~\cite{liu2023interngpt,shen2023hugginggpt,wu2023visual} utilize prompt engineering to connect independent modules with LLM API. 
Numerous works have attempted to align the visual and language modalities using various techniques, such as cross-attention mechanism~\cite{alayrac2022flamingo}, Q-Former~\cite{dai2023instructblipgeneralpurposevisionlanguagemodels,li2023blip}, prompt tuning~\cite{zhang2023llama}, linear projection~\cite{koh2023grounding,liu2023visual}, and unified model architectures~\cite{peng2023kosmos,xiao2024florence}. 

The unprecedented reasoning ability of MLLM has been examined on diverse vision-language tasks~\cite{lv2023kosmos,wang2025cogvlm}. By connecting with a suitable vision decoder, MLLM can also be applied to traditional vision-centric tasks, such as image segmentation~\cite{lai2024lisa,xia2024gsva,zhang2023next,he2024multi,bao2024cores}, object detection~\cite{zhang2023next,zang2024contextual,you2023ferret,chen2023shikra,ma2024groma}, object tracking~\cite{wang2024elysium,munasinghe2023pg}, etc. Among them, LISA~\cite{lai2024lisa} demonstrates the strongest reasoning ability in reasoning segmentation, where a [SEG] token is introduced to guide the mask decoder that receives latent features from a pre-trained visual backbone. However, LISA and its follow-up work are still prone to errors when tackling complex and challenging images or unclear or highly implicit text descriptions. 

\subsection{Image Segmentation} 
Semantic segmentation is an important computer vision task that assigns a class label to every pixel of a given image. Early methods primarily relied on  graphical models like conditional random fields (CRFs)~\cite{chen2017deeplab,krahenbuhl2011efficient} and region growing~\cite{dias2019semantic}. Over the past decade, various model architectures have been proposed to effectively extract the semantic information and improve the state-of-the-art performance, such as encoder-decoder structure~\cite{badrinarayanan2017segnet}, dilated
convolution~\cite{yu2015multi}, pyramid pooling module~\cite{zhao2017pyramid}, and non-local operator~\cite{liu2015parsenet}. Follow-up works on instance segmentation~\cite{cheng2022masked,he2017mask} and panoptic
segmentation~\cite{cheng2020panoptic,kirillov2019panoptic} further advanced the instance-level image segmentation.

Recently, the Segment Anything Model (SAM)~\cite{kirillov2023segment} has revolutionized image segmentation. By training with billions of images and masks, SAM serves as a foundation segmentation model, enabling promptable segmentation with multimodal inputs like points and bounding boxes. This inspires researchers to contribute to numerous works of reasoning segmentation. By incorporating SAM with multimodal LLM, subsequent works~\cite{lai2024lisa,xia2024gsva,zhang2023next,he2024multi} showcase strong capabilities of generating the mask with respect to a complex or implicit text description. However, while SAM has shown outstanding generalization ability in diverse scenarios, it still often fails in challenging images with intricate structures, blurry boundaries, occlusions, or high similarity with surroundings~\cite{fan2020camouflaged} without careful or additional prompt considerations. By leveraging chain of thoughts from MLLMs,   we found that the robustness of SAM can be substantially improved especially  in challenging cases. Specifically, we propose to let LLM ``think'' autonomously before making decisions, instead of  directly generating segmentation tokens to interact with SAM. 

\subsection{Chain of Thoughts}
CoT aims at improving the reasoning performance of LLMs via a series of intermediate reasoning steps~\cite{wei2022chain,wang2022self}. Existing CoT methods typically focus on inference with progressive reasoning demonstrations~\cite{zhang2022automaticchainthoughtprompting,wei2022chain,lyu2023faithful} or explicit step-by-step prompts~\cite{kojima2023largelanguagemodelszeroshot}. However, in the context of multimodal LLMs, these approaches often fail to yield significant improvements due to model capacity constraints. Multimodal LLMs usually acquire CoT capabilities through fine-tuning on specially constructed multimodal CoT datasets~\cite{mondal2024kamcotknowledgeaugmentedmultimodal,zhang2024multimodalchainofthoughtreasoninglanguage,lu2022learn} or integrating complex intermediate representations like graphs~\cite{mitra2024compositional} and code~\cite{suris2023vipergpt}, thereby limiting their accessibility and scalability. In this work, we aim to bypasses these limitations by approaching training-free CoT prompts of multimodal LLM for reasoning segmentation. 


Multimodal LLMs with CoT are recently integrated for reasoning segmentation. CoReS~\cite{bao2024cores} attempts to solve reasoning segmentation with CoT, while the MLLM framework fails to generate a detailed chain of thoughts for deep comprehension. Rather, it focused on decomposing the segmentation problem into two successive tasks: localization and mask decoding. Therefore, it still requires a detailed user text prompt to achieve high segmentation quality. A concurrent work Seg-Zero~\cite{liu2025seg} utilizes a reinforcement learning-based approach to train the reasoning model. While achieving impressive reasoning capability, Seg-zero requires a long training process and an intermediate bounding box, which limits the model from segmenting multiple objects within an image. In contrast, our proposed ThinkFirst is training-free, utilizing zero-shot understanding and reasoning capability from GPT-4o.
\section{Method}
\begin{figure*}
    \centering
    \includegraphics[width=.95\linewidth]{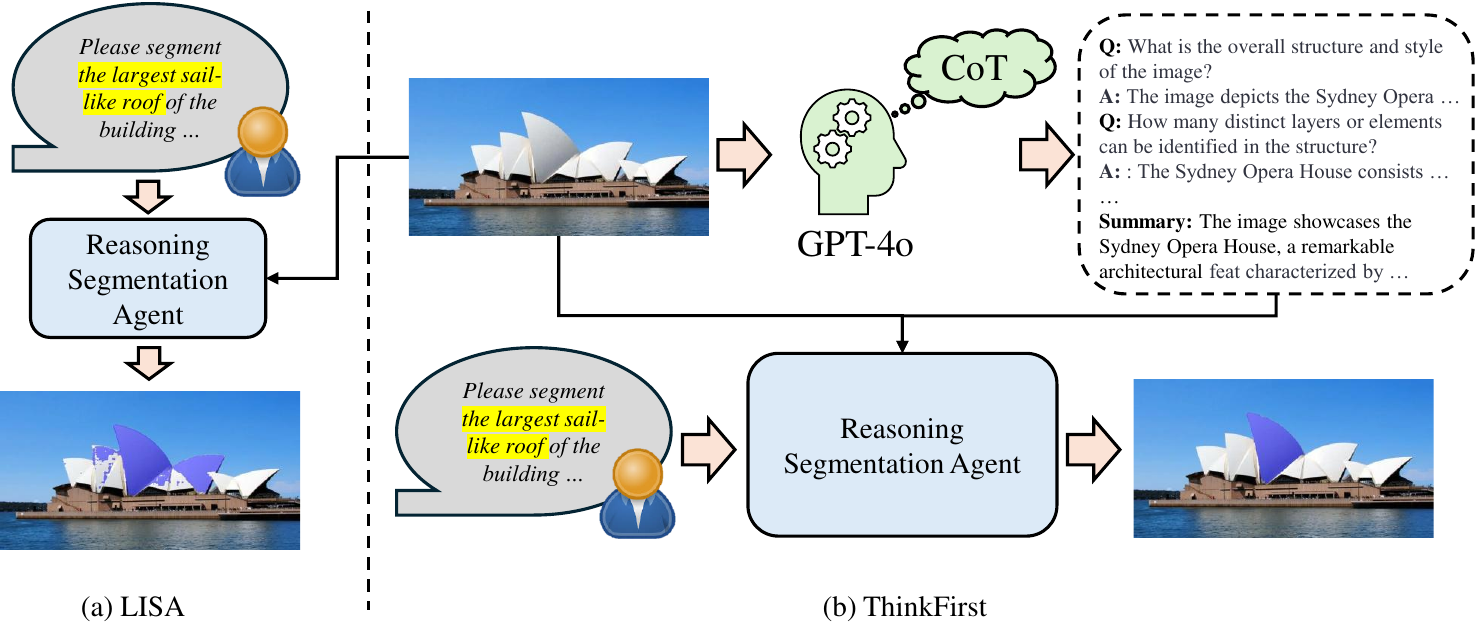}
    \vspace{-0.1in}
    \caption{\textbf{Overview.} Given an input image and a user query, our proposed ThinkFirst will first send the image to GPT-4o for chain-of-thought (CoT) information, which is later used to guide the inference of a reasoning segmentation agent, greatly improving the quality of reasoning segmentation.}
    \label{fig:overview}
    \vspace{-1em}
\end{figure*}
This section elaborates the detailed design of ThinkFirst, overviewed in Figure~\ref{fig:overview}.  We begin by introducing the preliminary knowledge of a reasoning segmentation agent in Section~\ref{method:preliminary}. Next we will detail how we leverage the GPT-4o's chain of thoughts in Section~\ref{method:thinkfirst} to improve the performance of reasoning segmentation. Following that, we provide the prompt details of ThinkFirst in Section~\ref{method:prompt}. Lastly, we will show in Section~\ref{method:control} that, by simply adjusting the GPT-4o prompt, ThinkFirst can process casual user's image-based control, such as scribbles, points, and bounding boxes.

\subsection{Preliminary: Reasoning Segmentation Agent}~\label{method:preliminary} 
Given an image $I$ and a textual query $q$, a reasoning segmentation agent aims to output the binary mask $\hat{\mathrm{M}}(I,q)$ corresponding to the best object instances that $q$ refers to. 

A reasoning segmentation agent usually consists of three modules: a vision encoder $E$, a mask decoder $\mathcal{D}$, and a vision-language model $\mathcal{F}$. During each inference, the image $I$ will be first encoded by the visual encoder and generate a visual-based feature $\mathbf{f}$. 
Simultaneously, the image $I$ is combined with the prompt $q$ and fed into the model $\mathcal{F}$ for a segmentation-purposed representation $\mathbf{h}$. Finally, the visual-based feature $\mathbf{f}$ and the segmentation-purposed representation $\mathbf{h}$ are fed to the mask decoder to generate outputs via a series of attention-based operations. The entire process can be formulated as:
\begin{gather}
    \mathbf{f}=E(I),\\
    \mathbf{h}=\mathcal{F}(I,q),\\
    \hat{\mathrm{M}}=\mathcal{D}(\mathbf{h},\mathbf{f}).
\end{gather}

In practice, the vision encoder $E$ and mask decoder 
$\mathcal{D}$ usually inherit the same architecture from SAM~\cite{zou2023segment}, where the vision encoder is frozen and the mask decoder requires further fine-tuning. The vision-language model $\mathcal{F}$ is fine-tuned with LoRA~\cite{hu2022lora}.

\subsection{ThinkFirst Overall Framework}~\label{method:thinkfirst}
Instead of directly feeding the input image $I$ and the user prompt $q$ to the reasoning segmentation agent, we first utilize GPT-4o~\cite{hurst2024gpt} for CoT pre-processing. The prompts for GPT-4o consists of two parts: environment prompt $P_e$ and task-specific prompt $P_t$, where environment prompt is universal and task-specific prompt varies by the problem settings. These prompts are combined and fed to GPT-4o for a long chain of thoughts $\{(Q_i,A_i)\}_{i=1}^n$. In specific, for the first pair of question and answer, the generative process can be formulated as
\begin{equation}
\begin{split}
    Q_1 &= \mathrm{GPT}(P_e,P_t,I),\\
    A_1 &= \mathrm{GPT}(P_e,P_t,I|Q_1).
\end{split}
\end{equation}
The follow-up chain of thoughts can be formulated as
\begin{equation}
\begin{split}
    Q_k &= \mathrm{GPT}(P_e,P_t,I|Q_1,A_1,...,Q_{k-1},A_{k-1}),\\
    A_k &= \mathrm{GPT}(P_e,P_t,I|Q_1,A_1,...,Q_{k-1},A_{k-1},Q_k),
\end{split}
\end{equation}
where $k\in\{2,3,...,n\}$. Notably, the actual value of $n$ is sampled from a probabilistic distribution. After generating the chain of thoughts, GPT-4o will generate a short paragraph of informative summary $S$. This process can be written as
\begin{equation}
    S=\mathrm{GPT}\left(P_e,P_t,I|Q_1,A_1,...,Q_n,A_n\right)
\end{equation}
Finally, we concatenate the summary $S$ with the user query $q$, generating the predictive binary mask $\hat{\mathrm{M}}(I,S\oplus q)$ with reasoning segmentation agent such as LISA~\cite{lai2024lisa}.

\subsection{ThinkFirst Prompt Details}~\label{method:prompt}

\noindent{\bf GPT-4o setup.} To maintain the stability of ThinkFirst among all inferences, we set the temperature of GPT-4o to 0.5, and use the following context \textit{``You are a helpful assistant that answers question as simple as possible.''}. We limit the maximal number of output tokens to 2,000, such that GPT-4o does not generate redundant descriptions. 

\vspace{0.5em}
\noindent{\bf Environment prompt.} The environment prompt $P_e$ aims to guide the behaviors of the GPT-4o. We use the following prompt to set up the environment for GPT-4o:

\vspace{-0.1em}
{\begin{leftbar}
\noindent{\small\textit{``You will serve as an agent for language-based image segmentation model. During each inference, your task is to describe a given image with chain of thoughts. You need to provide as many details as possible to help the segmentation model understand the image better. The target objects may contain multiple layers, be blocked by other objects, or be seamlessly embedded in their surroundings. Your description will be later sent to the segmentation as prompt. For example, if given an image, you need to describe what can be seen in the image, the number of objects for each categories, the position of each object, the structure of the object, the number of layers of the object, etc. The actual description depends on the given image.}

\noindent{\small\textit{For the output, you need to follow the format:}}

\small\textit{- \texttt{<}Question 1\texttt{>}: \texttt{<}Answer 1\texttt{>}.}

\small\textit{- \texttt{<}Question 2\texttt{>}: \texttt{<}Answer 2\texttt{>}.}

\noindent{\small\textit{..., etc, where each pair of prompt and answer implies the chain of thoughts, i.e., different levels or different part of the image understanding. For example, the first prompt can be related to the overall style or background of the image. Finally, you need to summarize the description based on your generated prompts and answers with the format:}}

\small\textit{- Summary: \texttt{<}your summary\texttt{>}}.''}
\end{leftbar}}
\vspace{-0.1em}

\begin{figure*}
    \centering
    \includegraphics[width=\linewidth]{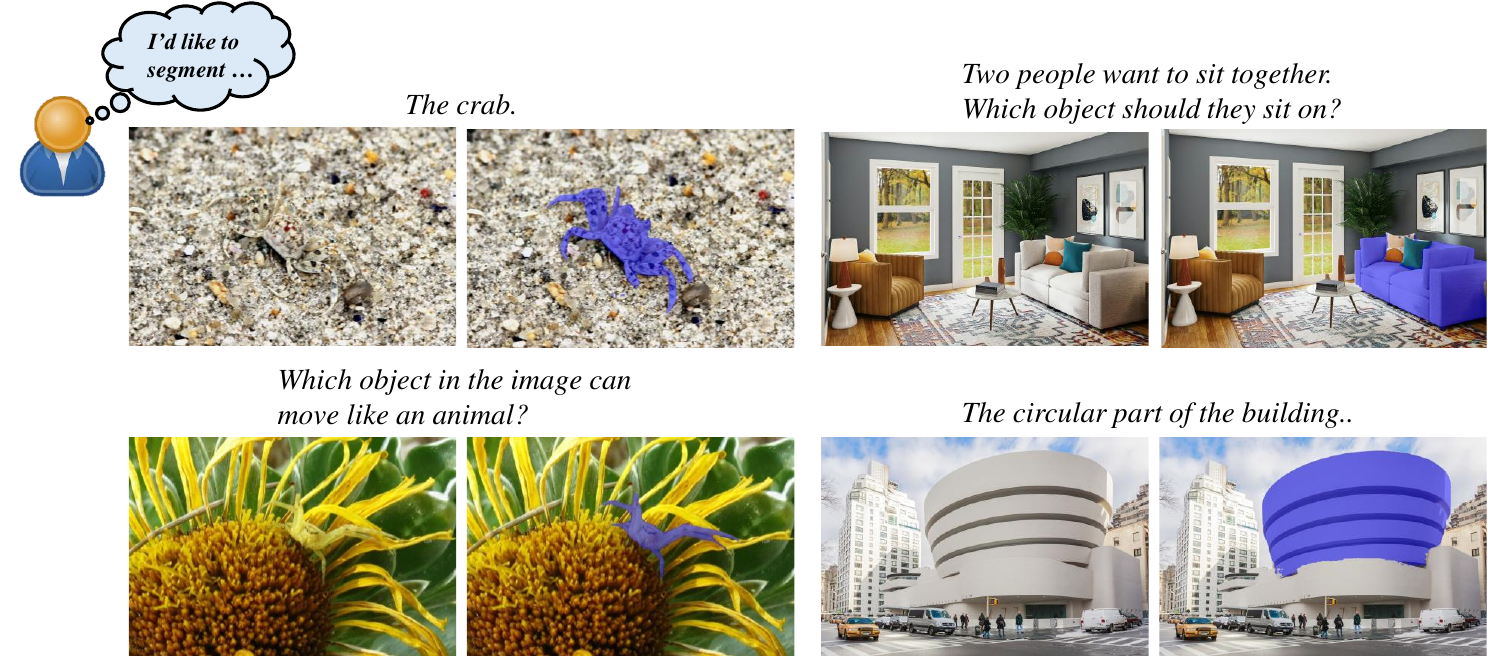}
    \vspace{-0.2in}
    \caption{Complex results. Refer to the appendix for additional complex results.}
    \vspace{-0.1in}
    \label{fig:complicate}
\end{figure*}

\noindent\textbf{Task-specific prompt.}
The task-specific prompt $P_t$ aims to provide simple and high-level instructions for GPT-4o, which are sent along with the image. In simple cases, we formulate it as \textit{``Please describe the image.''} In more difficult cases, users may provide additional information for specific goals. For example, we use the task-specific prompt: \textit{``Please describe the image and find the camouflaged objects if any.''} when tackling camouflaged object detection with implicit queries.

\subsection{Casual Image-based Controls}~\label{method:control} 
ThinkFirst can process diverse image-based controls in addition to text, such as points, bounding boxes, scribbles casually drawon on query images. This input setting is more flexible than SAM~\cite{zou2023segment} and allows the user to further refine the object of interests after text-based segmentation. Given an image $I$, a control image $I_{ann}$ with annotations, and a query $q$. We first define the task-specific prompt $P_t$ as follows:

\begin{leftbar}
\noindent{\small\textit{``Note that your task is slightly different for this input image. A particular part of the object is manually annotated with a circle, a star point, or a bounding box. Please describe the image with chain of thoughts and identify which part it is. Your summary needs to include the description for the target part, so that the segmentation model know how to output the mask. After summarizing the description, please generate a prompt in the format of `The image features a .... Please segment the xxx in the image.', where xxx implies what the annotated part is. Do not mention what the annotation type is in the prompt.''}}
\end{leftbar}

Then, we feed $I_{ann}, P_e, P_t$ to GPT-4o, where $P_e$ is identical as mentioned in Section~\ref{method:prompt}. This will generate a chain of thoughts $\{(Q_i,A_i)\}_{i=1}^n$, a summary $S$, and an output prompt $\widetilde{P}$. Finally, we use $\widetilde{P}$ to segment the original image $I$ for the output mask $\hat{\mathrm{M}}(I,\widetilde{P})$ with reasoning segmentation agent to achieve  segmentation with casual cues marked on the image.


\section{Experiments}
In this section we empirically validate the effectiveness of our proposed ThinkFirst framework on solving reasoning segmentation problems. We start by introducing the experimental settings in Section~\ref{exp:settings}. Then, we show the qualitative and quantitative results in Section~\ref{exp:results} and provide insightful analysis. Next, we will proceed to introduce an interesting application.
Section~\ref{exp:waldo} applies ThinkFirst to solve a classic game known as \textit{``Where's Waldo''} which pushes a reasoning segmentation model under testing to  limits when we humans may not be able to spot Waldo without effort.  Finally, we will conduct ablation studies in Section~\ref{exp:ablation} to show the necessity of GPT-4o CoT reasoning.
\begin{figure}
    \centering
    \includegraphics[width=\linewidth]{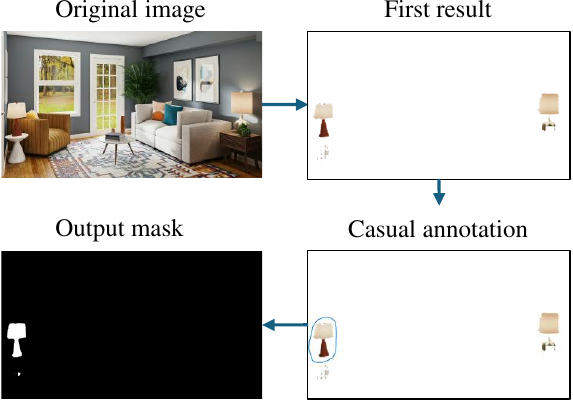}
    \vspace{-0.2in}
    \caption{Casual image-based control can be easily used to refine the reasoning segmentation result. We use the query ``Please segment the lamp next to the armchair.'' in this example, where the mask is refined with a hand-drawn circle.}
    \vspace{-0.15in}
    \label{fig:refine}
\end{figure}
\begin{table*}[]
    \centering
    \caption{Quantitative evaluation on camouflaged images in \textit{COD-10K}~\cite{fan2020camouflaged}. Note that the models are blind to both dataset divisions.}
    \resizebox{.85\linewidth}{!}{%
    \begin{tabular}{l|cc|cc|cc|cc}
    \toprule
        &\multicolumn{4}{c|}{Train set}&\multicolumn{4}{c}{Test set}\\
        Method&\multicolumn{2}{c|}{Implicit query}&\multicolumn{2}{c|}{Explicit query}&\multicolumn{2}{c|}{Implicit query}&\multicolumn{2}{c}{Explicit query}\\\cmidrule(lr){2-9}
         &gIoU&cIoU&gIoU&cIoU&gIoU&cIoU&gIoU&cIoU\\ \midrule \midrule
         LISA-13B-Llama2~\cite{lai2024lisa}&65.0&60.3&66.5&59.7&63.4&55.8&\textbf{66.5}&58.0\\
         ThinkFirst-LISA-13B-Llama2&\textbf{68.0}&\textbf{70.6}&\textbf{67.2}&\textbf{68.9}&\textbf{65.6}&\textbf{65.0}&66.1&\textbf{68.7}\\
         
         \bottomrule
    \end{tabular}
    }
    \vspace{-0.1in}
    \label{tab:cod}
\end{table*}

\vspace{-0.1in}
\subsection{Experimental Settings}~\label{exp:settings}

\vspace{-0.05in}
\noindent{\bf Dataset.} We focus on evaluating ThinkFirst on camouflaged images in COD10K~\cite{fan2020camouflaged} and underwater images in MVK~\cite{MVK}. The COD10K dataset contains 69 camouflaged classes and 5,066 camouflaged images (3,040 in train set and 2,026 in test set), where objects are ``seamlessly'' embedded in their surroundings. The MVK dataset is a video dataset for underwater scenes, where we sample random frames for qualitative evaluation. Additionally, we use the ReasonSeg~\cite{lai2024lisa} dataset as evaluation for standard images without special distortion or coloring conditions, where each image is queried with three different user prompts.

\vspace{0.5em}
\noindent{\bf Task.} We use both the explicit and implicit queries to define $q$ for ThinkFirst evaluation. The explicit query follows the form \textit{``Please segment the }\texttt{<OBJ>}\textit{ in the image.''}, where \texttt{<OBJ>} refers to the object class. The implicit query is \textit{``What is the camouflaged object in the image that can move like an animal? Please segment it.''}. 

\vspace{0.5em}
\noindent{\bf Model.} We select the open-sourced \textit{LISA-13B-Llama2} as our reasoning segmentation backbone. When testing the ThinkFirst results, we use fp16 with 8-bit quantization to prevent out-of-memory issues, which may slightly harm the performance compared to the full model.

\vspace{0.5em}
\noindent{\bf Evaluation metric.} We follow the previous works of reasoning segmentation and use cIoU and gIoU to measure the segmentation quality. gIoU is defined by the average of all per-image Intersection-over-Unions (IoUs), while cIoU is defined by the cumulative intersection over the cumulative union.

\begin{figure*}
    \centering
    \includegraphics[width=\linewidth]{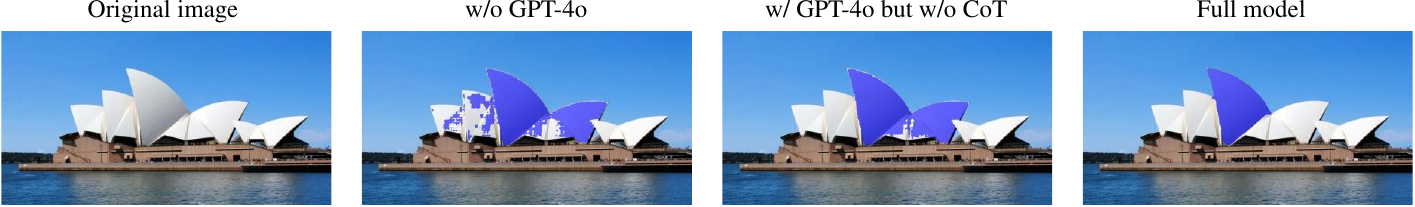}
    \vspace{-0.2in}
    \caption{\textbf{Ablation study.} We examine how the chain of thoughts from GPT-4o improve the reasoning segmentation. This figure shows the results with user query \textit{``Please segment the largest sail-like roof of the building.''}.}
    \vspace{-0.17in}
    \label{fig:ablation}
\end{figure*}
\begin{figure}
    \centering
    \includegraphics[width=\linewidth]{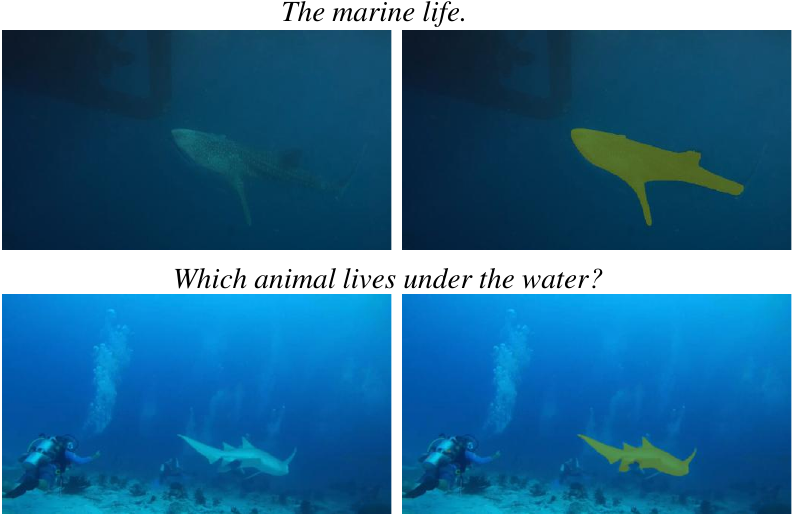}
    \vspace{-0.25in}
    \caption{Qualitative results of ThinkFirst for underwater images from the MVK dataset~\cite{MVK}, where objects are captured under severe blurry and color shift condition.}
    \label{fig:marine}
    \vspace{-0.12in}
\end{figure}

\subsection{Main Results}~\label{exp:results}
\hspace{-0.17in}\textbf{Quantitative evaluation.} The quantitative results of COD10K are presented in Table~\ref{tab:cod}. For both the explicit and implicit queries, our ThinkFirst framework surpasses the baseline by a clear margin in both dataset divisions. When processing implicit queries, ThinkFirst outperforms the baselines by $2.2-3.0\%$ gIoU and $9.2-10.3\%$ cIoU. When dealing with explicit queries, ThinkFirst also improves $9.2-10.7\%$ cIoU in comparison with LISA. This exhibits the exceptional capability of ThinkFirst for image understanding and reasoning. We also show the quantitative results of ReasonSeg in Table~\ref{tab:reasonseg}, where ThinkFirst achieves comparable segmentation quality with previous works on standard images.

\vspace{0.5em}
\noindent{\bf Qualitative evaluation.} Figure~\ref{fig:qual}
demonstrates qualitative comparison of ThinkFirst with the baseline on COD10K, 
where ThinkFirst is clearly a winner in segmenting camouflaged objects in difficult scenarios. Also, more complex examples, such as out-of-domain objects, highly implicit user queries, and underwater images are showcased in Figure~\ref{fig:complicate} and \ref{fig:marine} for evaluation of the ThinkFirst reasoning capability. More examples can be found in the appendix.

\begin{figure}
    \centering
    \includegraphics[width=\linewidth]{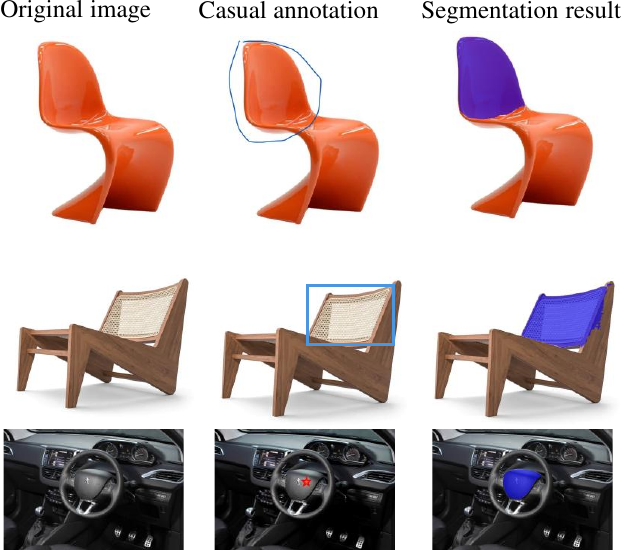}
    \vspace{-0.2in}
    \caption{Part segmentation with ThinkFirst (chair's backrest and horn on steering wheel), which enables reasoning segmentation with user's casual annotations, such as scribbles, bounding boxes, or star points. See more examples in the appendix.}
    \label{fig:scribble}
    \vspace{-0.15in}
\end{figure}

\vspace{0.5em}
\noindent{\bf Casual image-based control.} Figure~\ref{fig:scribble} showcases the examples of segmentation with casual image-based controls, including  a manual scribble, a bounding box, and a star point, which significantly increasing the user accessibility of ThinkFirst in real practice. These controls can also be used to manually refine the segmentation result, as shown in Figure~\ref{fig:refine}.

\begin{figure}
    \centering
    \includegraphics[width=\linewidth]{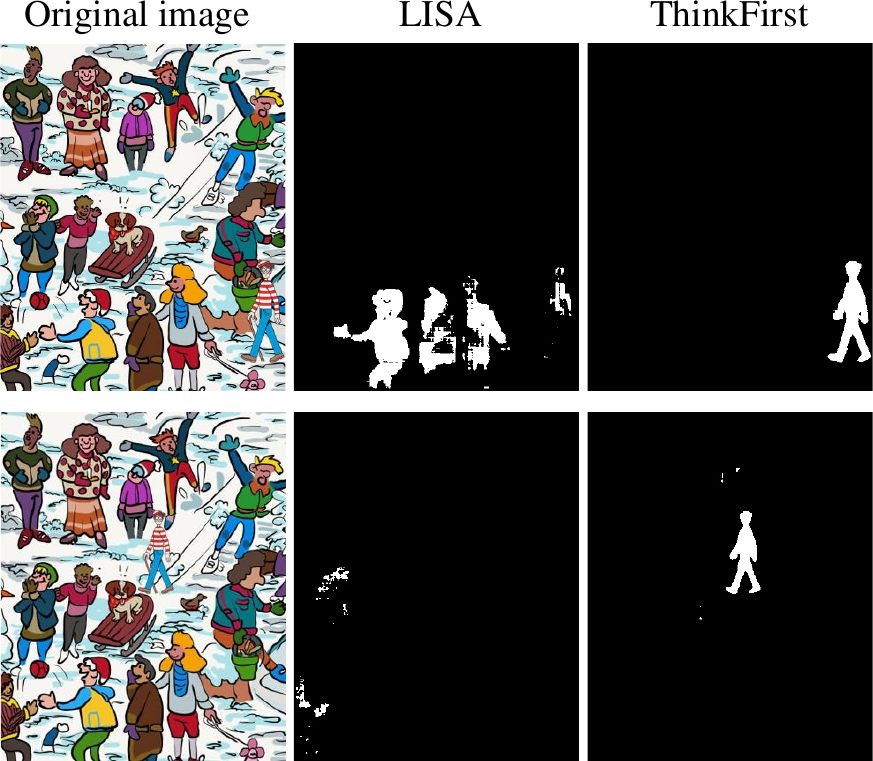}
    \vspace{-0.2in}
    \caption{\textbf{Where is Waldo?} For LISA baseline~\cite{lai2024lisa}, we use the query \textit{``Please segment Waldo, a boy wearing red and white striped shirt.''}. For our ThinkFirst-LISA method, we let GPT-4o to directly generate a pseudo-query to prompt LISA.}
    \label{fig:waldo}
\end{figure}

\subsection{Application: Where is Waldo?}~\label{exp:waldo}
\textit{``Where's Waldo?''}, also referred to as \textit{Where's Wally?}, is a visual search puzzle game that challenges players to locate a character, Waldo, among densely populated and intricately detailed illustrations, where Waldo is a boy in a red and white striped shirt. Previous methods generally struggle to find the location of Waldo due to the complex scenes, i.e., there is no obvious foreground and focus in the image.

We ask GPT-4o to think with chains of thoughts and eventually find the location of Waldo by slightly adjusting the environment and task-specific prompts. For this game-playing application, please refer to the prompting details in the appendix: Given a game image $I$ of Where's Waldo, we follow ThinkFirst's pipeline to let GPT-4o generate a pseudo-user query $\widetilde{P}$ based on $I$. This query is then fed to the reasoning segmentation for the mask $\hat{\mathrm{M}}(I,\widetilde{P})$. As shown in Figure~\ref{fig:waldo}, ThinkFirst is able to find Waldo among numerous characters in a complex scene, while simply using the reasoning segmentation model fails to do so.

\subsection{Ablation Studies}~\label{exp:ablation}
\hspace{-0.1in}\textbf{Chain of thoughts.} We conduct experiments to verify the effectiveness of each component in ThinkFirst. Specifically, we compare ThinkFirst with two models. The first one, referred as ``w/o GPT-4o'', generates the mask directly from the image and the user query. Another model, i.e. ``w/ GPT-4o but w/o CoT'', guides the reasoning segmentation agent with GPT-4o output, while simply using the prompt ``Please describe the image'' for GPT-4o without using CoT reasoning. As shown in Figure~\ref{fig:ablation}, GPT-4o's guidance slightly improve the segmentation quality but fail to focus on the image details. Instead, using the full model of ThinkFirst results in the best performance.

\vspace{0.5em}
\noindent{\bf GPT-4o output.} We show the questions and summary from a sampled chain of thoughts of GPT-4o in the ThinkFirst framework for reference. Refer to the full output in the appendix.

\begin{table}[]
\centering
\caption{Quantitative evaluation on the test set of \textit{ReasonSeg}~\cite{lai2024lisa}. Baselines are excerpted from~\cite{lai2024lisa}.}
\vspace{-0.1in}
\setlength{\tabcolsep}{5.5mm}{
\renewcommand\arraystretch{1.0}
\resizebox{.85\linewidth}{!}{%
\begin{tabular}{l|c c}
\toprule
\multirow{2}*{Method}&\multicolumn{2}{c}{ReasonSeg (overall)}\\
&gIoU&cIoU\\\midrule\midrule
OVSeg~\cite{liang2023open}&26.1&20.1\\
GRES~\cite{liu2023gres}&21.3&22.0\\
X-Decoder~\cite{zou2023generalized}&21.7&16.3\\
SEEM~\cite{zou2023segment}& 24.3&18.7\\
LISA-13B~\cite{lai2024lisa} &44.8&45.8\\
ThinkFirst&\textbf{48.6}&\textbf{49.2}\\

\bottomrule
\end{tabular}
\label{tab:reasonseg}
}
}
\end{table}
\begin{center}

\vspace{0.3em}
\noindent\fbox{\begin{minipage}[t]{0.95\linewidth}
\begin{wrapfigure}{r}{.5\linewidth}
\includegraphics[width=\linewidth]{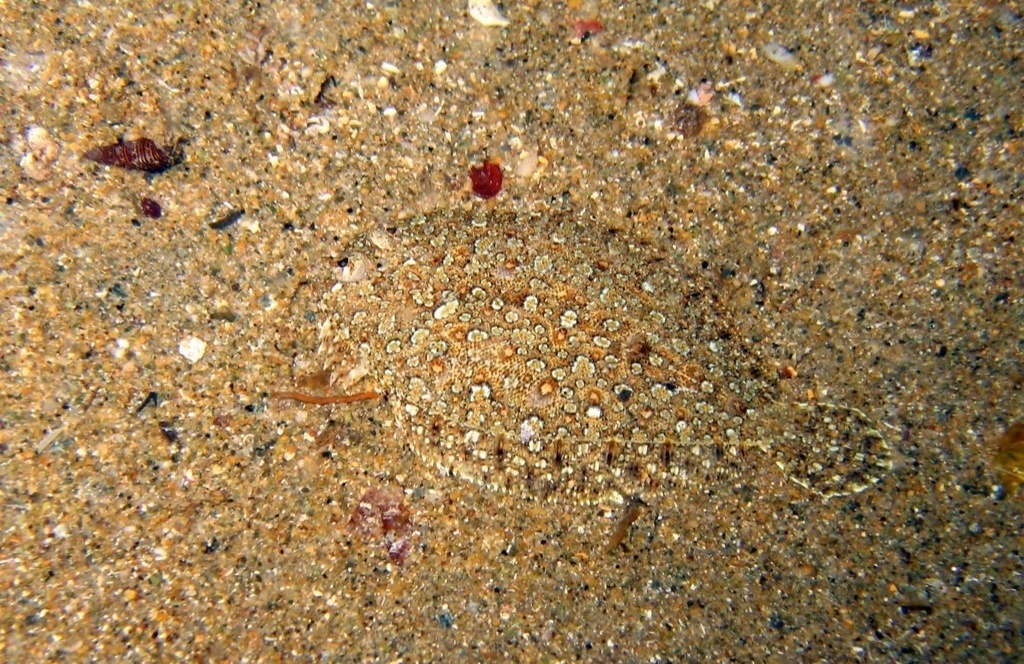}
\caption{Sample input image for GPT-4o.}
\end{wrapfigure} 
\noindent\small{\textbf{Q1:} What is the overall setting of the image?}

\noindent\small{\textbf{Q2:} What objects are present in the image?}

\noindent\small{\textbf{Q3:} How is the camouflaged object positioned in relation to its surroundings?
}

\noindent\small{\textbf{Q4:} Describe the structure and appearance of the camouflaged object.}

\noindent\small{\textbf{Q5:} Are there any layers or additional features to note about the objects?}

\noindent\small{\textbf{Q6:} Is there any other notable detail in the image?}

\noindent\small{\textbf{Summary:} The image showcases an underwater sandy environment where a camouflaged flatfish is prominently positioned towards the center-right, blending into the sandy substrate with its mottled coloration. The flatfish exhibits a flattened, oval shape and is partially embedded in the sand, making it difficult to distinguish from its surroundings. A small shell is also present to the left of the fish, contributing to the complexity of the scene. The overall setting is rich in texture, with various organic materials scattered across the sand.
}
\end{minipage}} 
\end{center}

\section{Discussion}~\label{sec:conclusion}
\begin{table}
\begin{center}
\caption{Referring expression segmentation results on RefCOCO~\cite{kazemzadeh2014referitgame} dataset. The cIoU metrics of each split are reported. Baselines are excerpted from~\cite{lai2024lisa}. 
}
\vspace{-0.1in}
\setlength{\tabcolsep}{5.5mm}{
\renewcommand\arraystretch{1.0}
\resizebox{.9\linewidth}{!}{
\begin{tabular}{l|ccc}
\toprule
Method& Val. & Test-A & Test-B \\
\midrule
\midrule
MCN~\cite{luo2020multi} & 62.4 & 64.2 & 59.7 \\
VLT~\cite{ding2021vision} & 67.5 & 70.5 & 65.2 \\
CRIS~\cite{wang2022cris} & 70.5 & 73.2 & 66.1 \\
LAVT~\cite{yang2022lavt} & 72.7 & 75.8 & {68.8} \\
\midrule
LISA~\cite{lai2024lisa} & \textbf{73.4} & 76.2 & \textbf{69.5} \\
ThinkFirst&70.4&\textbf{76.3}&64.1\\
\bottomrule
\end{tabular}}
\label{tab:refer}
}
\end{center}
\vspace{-0.2in}
\end{table}

\hspace{-0.11in}{\bf Limitations.} ThinkFirst shows little improvement, while still being comparable, when testing on traditional datasets such as RefCOCO~\cite{kazemzadeh2014referitgame}, as shown in Table~\ref{tab:refer}. We do not use the three common datasets (RefCOCO, RefCOCO+~\cite{kazemzadeh2014referitgame}, and RefCOCOg~\cite{mao2016generation}) for evaluation due to the potential data leakage issue. Specifically, segmentation models like LISA~\cite{lai2024lisa} are often pretrained on the MSCOCO~\cite{lin2014microsoft} dataset, while these three datasets are all subsets of the train division in MSCOCO. To uphold fairness of comparison, we mainly tested ThinkFirst on out-of-domain datasets with a substantial difference from MSCOCO. To enhance the capability for ThinkFirst on traditional datasets requires further exploration of other GPT-4o prompts.

\vspace{0.5em}
\noindent{\bf Concluding remarks} This paper introduces ThinkFirst, a novel framework for reasoning segmentation that leverages the chain-of-thought (CoT) capabilities of GPT-4o to enhance the robustness of reasoning segmentation outputs. Our approach demonstrates significant improvements over existing methods, particularly in challenging scenarios involving camouflaged objects, ambiguous boundaries, and out-of-domain instances. Besides the exceptional reasoning capability, ThinkFirst accepts multimodal image controls, thereby enhancing accessibility and usability for a broader audience. We validate the effectiveness of our framework across diverse datasets, including COD10K~\cite{fan2020camouflaged}, MVK~\cite{MVK}, and the ``Where's Waldo?" challenge, both qualitatively and quantitatively. The results indicate that our CoT-based approach not only improves segmentation quality but also makes it a powerful tool for visual understanding tasks. Future work will explore refinements to the framework and the applications in other vision-language challenges using GPT-4o CoT.
\clearpage
\appendix
\renewcommand{\thefigure}{A\arabic{figure}}
\renewcommand{\thetable}{A\arabic{table}}
\renewcommand{\theequation}{A\arabic{equation}}
\setcounter{figure}{0}
\setcounter{table}{0}
\setcounter{equation}{0}

\vspace{0.5em}
\begin{center}
    \bf \Large Appendix
\end{center}
\vspace{0.5em}

\section{More details}
\paragraph{Task-specific prompt.} We use $P_t=$\textit{``Please describe the image and find the camouflaged objects if any.''} for testing the implicit query on COD10K~\cite{fan2020camouflaged}. When using the explicit query, we set $P_t$ to \textit{``Please describe the image and find the }\texttt{<}\textit{object}\texttt{>}\textit{.''}, where \texttt{<}\textit{object}\texttt{>} corresponds to the image category like butterfly, crocodile, crab, etc. For experiments on MVK~\cite{MVK}, ReasonSeg~\cite{lai2024lisa} and RefCOCO~\cite{kazemzadeh2014referitgame}, we simply set $P_t$ to \textit{``Please describe the image.''} 
\paragraph{Prompt for Where's Waldo}
We use the following prompt ($P_e$ and $P_t$) for GPT-4o CoT to tackle the Where's Waldo challenge: 

\begin{leftbar}
    \noindent{\small$\mathbf{P_e}$\textbf{: }\textit{``You are going to play a game called `Where's Waldo'. You will serve as an agent for language-based image segmentation model. During each inference, your task is to describe a given image with chain of thoughts. You need to provide as many details as possible to help the segmentation model understand the image better. Waldo is a boy wearing red and white striped shirt. He is somewhere hidden in the image. He may or may not be partially blocked by other people or objects. Your description will be later sent to the segmentation as prompt. For example, if given an image, you need to describe the overview of the image, what can be seen in the image, where Waldo is, what Waldo is doing, what Waldo is wearing, etc. The actual description depends on the given image.}}
    
    \noindent{\small\textit{For the output, you need to follow the format:}}
    
    \small\textit{- \texttt{<}Question 1\texttt{>}: \texttt{<}Answer 1\texttt{>}.}
    
    \small\textit{- \texttt{<}Question 2\texttt{>}: \texttt{<}Answer 2\texttt{>}.}
    
    \noindent{\small\textit{..., etc, where each pair of prompt and answer implies the chain of thoughts, i.e., different levels or different part of the image understanding. For example, the first prompt can be related to the overall style or background of the image. After chain of thoughts, you need to summarize the description based on your generated prompts and answers with the format:}}
    
    \small\textit{- Summary: \texttt{<}your summary\texttt{>}.''}
\end{leftbar}

\begin{leftbar}
    \noindent{\small$\mathbf{P_t}$\textbf{: }\textit{``Your summary needs to be as short and accurate as possible, mainly focusing on the location and characteristics of Waldo. Finally, you are going to generate a prompt for the segmentation model. This prompt should follows the format:}}
    
    \small\textit{- Prompt: \texttt{<}your prompt\texttt{>}.} 
    
    \noindent{\small\textit{This prompt should start with the sentence `Please segment the boy ...', where the boy refers to Waldo. This prompt should include the detailed location and features of Waldo.}}
\end{leftbar}

\noindent After feeding $P_e$ and $P_t$ to GPT-4o, it will generate a pseudo-user prompt $\widetilde{P}$ at the end, which is used to query the reasoning segmentation agent for the estimated mask.

\paragraph{GPT-4o full CoT:}
We provide some GPT-4o full CoT outputs for reference:
\begin{center}
\noindent\fbox{\begin{minipage}[t]{0.95\linewidth}
\begin{wrapfigure}{r}{.5\linewidth}
\includegraphics[width=\linewidth]{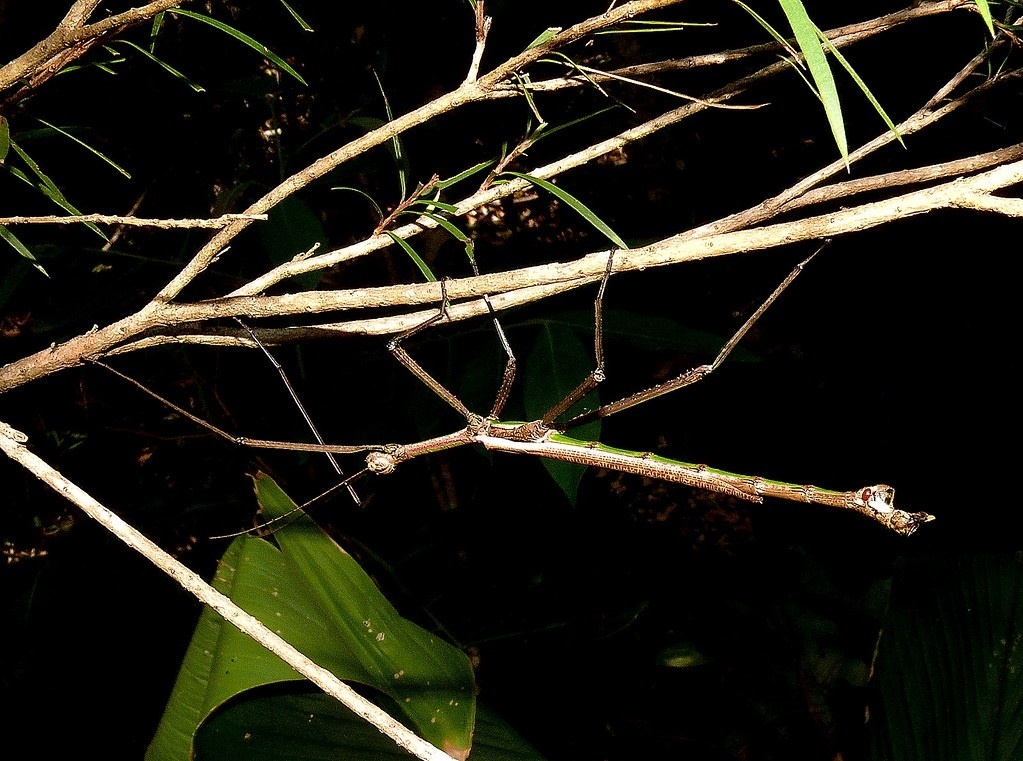}
\caption{Sample input image for GPT-4o. Refer to Figure~\ref{fig:qual} for the mask result.}
\end{wrapfigure} 
\noindent\small{\textbf{Q1:} What is the overall setting of the image?}

\noindent\small{\textbf{A1:} The image depicts a natural environment, specifically a dense area of vegetation, likely a forest or jungle. The background is dark, indicating that there is a lot of foliage blocking light, which contributes to a shadowy ambiance.}

\noindent\small{\textbf{Q2:} What objects are present in the image?}

\noindent\small{\textbf{A2:} The primary objects in the image are several thin branches and twigs, along with a camouflaged insect that resembles a stick or twig.}

\noindent\small{\textbf{Q3:} How many objects are there in each category?}

\noindent\small{\textbf{A3:} There are numerous branches and twigs in the background, but they are difficult to count precisely due to their overlapping nature. The camouflaged insect, which is a stick insect, appears to be a single individual.}

\noindent\small{\textbf{Q4:} Describe the position of the camouflaged object in relation to the surrounding objects.}

\noindent\small{\textbf{A4:} The camouflaged stick insect is positioned horizontally along a branch, blending in seamlessly with the twigs around it. Its body aligns closely with the branches, making it hard to distinguish from the surrounding environment.}

\noindent\small{\textbf{Q5:} What is the structure of the camouflaged object?}

\noindent\small{\textbf{A5:} The stick insect has a long, slender body that resembles a twig, with elongated legs that mimic the appearance of branches. Its coloration is predominantly brown and green, which helps it blend in with the natural colors of the branches.}

\noindent\small{\textbf{Q6:} Are there any layers or overlapping objects in the image?}

\noindent\small{\textbf{A6:} Yes, the branches create a layered effect, with some twigs overlapping others. The stick insect is partially obscured by the branches, enhancing its camouflage.}

\noindent\small{\textbf{Q7:} Is there any additional detail that enhances the camouflage of the object?}

\noindent\small{\textbf{A7:} The stick insect's texture and coloration closely match that of the twigs, with slight variations that mimic the bark and surface of the branches. Its posture, which is still and aligned with the branches, further aids in its concealment.}

\noindent\small{\textbf{Summary:} The image shows a dark, lush forest setting filled with overlapping branches and twigs. A single stick insect is camouflaged among the branches, positioned horizontally and blending in seamlessly with its surroundings. The insect's long, slender body and legs resemble twigs, and its coloration of brown and green enhances its camouflage. The overlapping nature of the branches adds complexity to the scene, making the insect difficult to spot.}
\end{minipage}} 
\end{center}
\begin{center}
\noindent\fbox{\begin{minipage}[t]{0.95\linewidth}
\begin{wrapfigure}{r}{.62\linewidth}
\includegraphics[width=\linewidth]{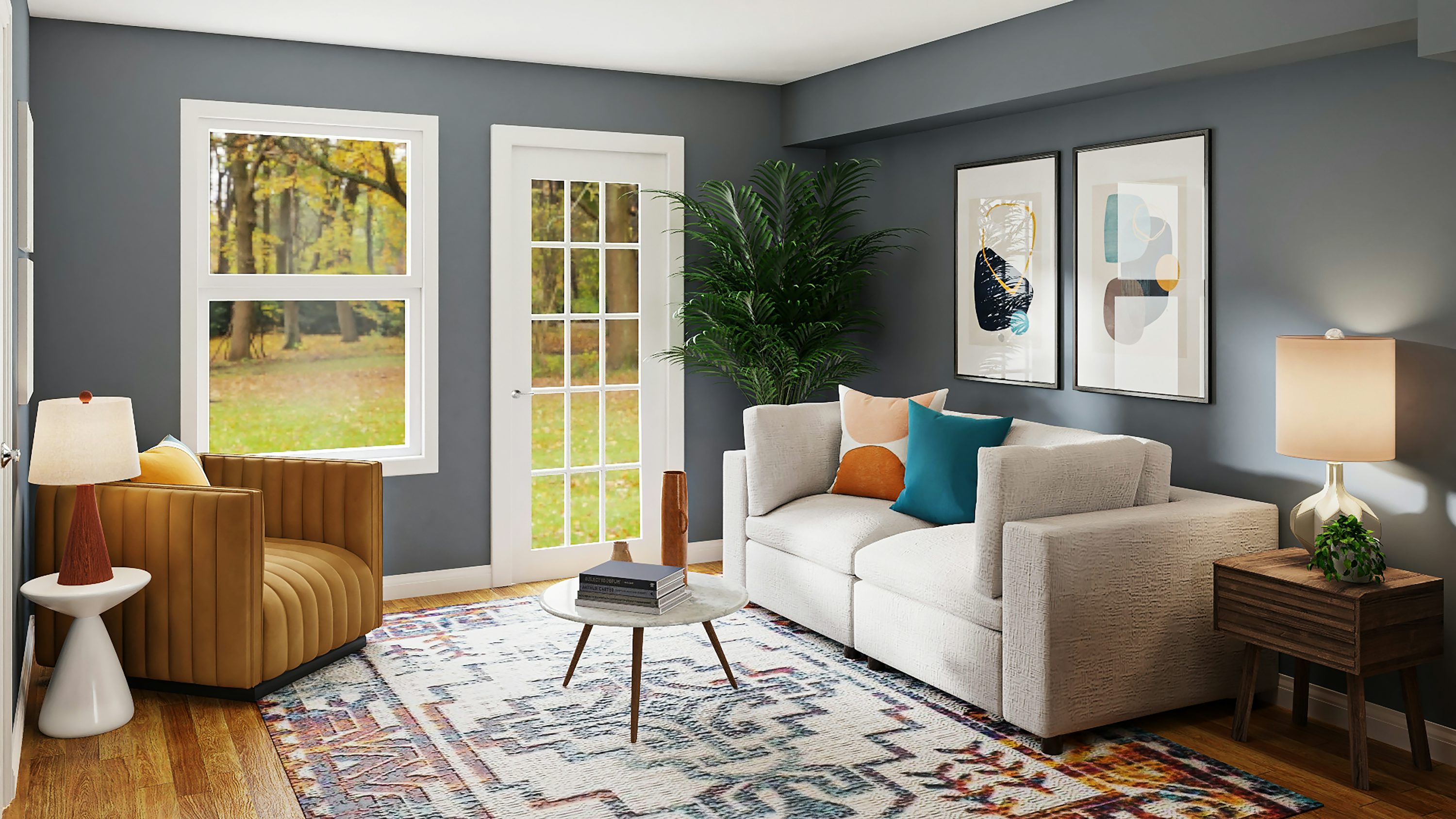}
\caption{Sample input image for GPT-4o. Refer to Figure~\ref{fig:complicate} and \ref{fig:refine} for the mask results.}
\end{wrapfigure} 
\noindent\small{\textbf{Q1:} What is the overall style or theme of the image?}

\noindent\small{\textbf{A1:} The image depicts a modern and cozy living room with a contemporary design. The color palette consists of soft neutrals, warm tones, and a touch of vibrant colors, creating a welcoming atmosphere.}

\noindent\small{\textbf{Q2:} How many main furniture pieces are present in the image, and what are their types?}

\noindent\small{\textbf{A2:} There are three main furniture pieces: a light-colored sofa, a mustard-colored accent chair, and a round coffee table. Additionally, there is a side table next to the sofa and a lamp.}

\noindent\small{\textbf{Q3:} What are the positions of the furniture pieces in the room?}

\noindent\small{\textbf{A3:} The sofa is positioned against the wall on the right side of the image, facing the center of the room. The accent chair is located in the bottom left corner near the window, and the coffee table is placed in front of the sofa, centrally located in the room. The side table with a lamp is next to the sofa on the right.}

\noindent\small{\textbf{Q4:} Describe the colors and textures of the furniture.}

\noindent\small{\textbf{A4:} The sofa is upholstered in a light, textured fabric, giving it a soft appearance. The accent chair is a rich mustard color with vertical ribbing for added texture. The coffee table has a marble top with wooden legs, while the side table is white with a smooth finish. The lamp has a glass base and a fabric shade.}

\noindent\small{\textbf{Q5:} Are there any decorative elements present in the room?}

\noindent\small{\textbf{A5:} Yes, there are several decorative elements including two framed art pieces on the wall above the sofa, which feature abstract designs in various colors. There is also a decorative plant beside the sofa, adding a touch of greenery to the space.}

\noindent\small{\textbf{Q6:} What can be seen outside the windows?}

\noindent\small{\textbf{A6:} Through the windows, there is a view of a natural landscape with trees that have autumn foliage, indicating the season. The sunlight filters through the leaves, creating a warm and inviting outdoor scene.}

\noindent\small{\textbf{Q7:} What type of flooring is visible in the image?}

\noindent\small{\textbf{A7:} The flooring is made of hardwood with a warm, natural finish, complementing the overall cozy aesthetic of the room. A patterned area rug is placed on top of the hardwood floor, adding color and texture.}

\noindent\small{\textbf{Summary:} The image showcases a modern living room characterized by a light-colored sofa, a mustard accent chair, and a round coffee table. The furniture is arranged to create an inviting atmosphere, with the sofa positioned against the wall and the accent chair in the corner. Decorative elements include abstract art on the walls and a plant beside the sofa, while the hardwood flooring and patterned rug enhance the cozy feel. Natural light streams in through the windows, revealing an autumn landscape outside.}
\end{minipage}} 
\end{center}
\begin{center}
\noindent\fbox{\begin{minipage}[t]{0.95\linewidth}
\begin{wrapfigure}{r}{.38\linewidth}
\includegraphics[width=\linewidth]{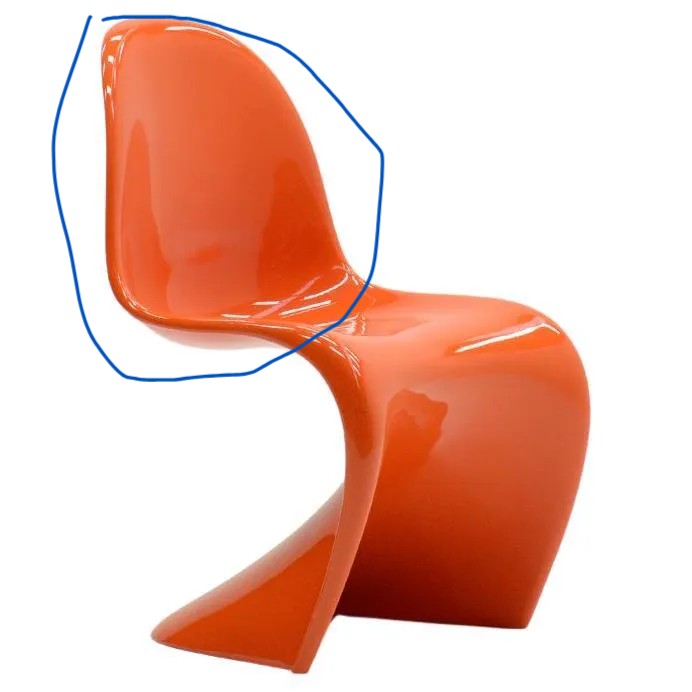}
\caption{Sample input image for GPT-4o. Refer to Section 3.4 for the details of image-based control and Figure~\ref{fig:scribble} for the mask results.}
\vspace{0.07in}
\end{wrapfigure} 
\noindent\small{\textbf{Q1:} What is the overall style and color of the object in the image?}

\noindent\small{\textbf{A1:} The image features a modern, minimalist chair with a sleek and curvy design. The chair is predominantly orange in color, giving it a vibrant and eye-catching appearance.}

\noindent\small{\textbf{Q2:} What are the main structural components of the chair?}

\noindent\small{\textbf{A2:} The chair consists of a single continuous piece that forms the seat, backrest, and legs. The backrest has a unique organic shape that flows into the seat, which is supported by a curved base that also serves as the legs.}

\noindent\small{\textbf{Q3:} How many distinct parts or layers can be identified in the chair?}

\noindent\small{\textbf{A3:} The chair appears to be a single-layer structure without any visible seams or joints, emphasizing its fluid design. However, the backrest has a slightly thicker profile compared to the seat, giving it a layered appearance from certain angles.}

\noindent\small{\textbf{Q4:} Where is the annotated part located, and what does it represent?}

\noindent\small{\textbf{A4:} The annotated part is located on the backrest of the chair, specifically highlighting the upper left section. This part represents the smooth, curved surface of the backrest, which transitions seamlessly into the seat.}

\noindent\small{\textbf{Q5:} What details can be observed about the texture and finish of the chair?}

\noindent\small{\textbf{A5:} The chair has a glossy finish that reflects light, enhancing its modern aesthetic. The surface appears smooth and shiny, contributing to its contemporary look.}

\noindent\small{\textbf{Q6:} Are there any other objects or background elements present in the image?}

\noindent\small{\textbf{A6:} The image is isolated, with no other objects or background elements visible. This focus on the chair allows for a clear view of its design and features.}

\noindent\small{\textbf{Summary:} The image features a modern, minimalist orange chair with a sleek, curvy design made from a single continuous piece. The chair consists of a seat, backrest, and legs, with the backrest having a slightly thicker profile. The annotated part highlights the upper left section of the backrest, showcasing its smooth, glossy surface. The chair stands alone in the image, emphasizing its distinctive design without any distractions.}

\noindent\small{\textbf{Prompt:} The image features a modern, minimalist chair with a sleek, curvy design. Please segment the upper left section of the backrest in the image.}
\end{minipage}} 
\end{center}

\newpage

\section{More results}
\paragraph{Camouflaged objects.} We provide more camouflaged examples from COD10K~\cite{fan2020camouflaged} in Figure~\ref{fig:more}.

\begin{figure*}
    \centering
    \includegraphics[width=\linewidth]{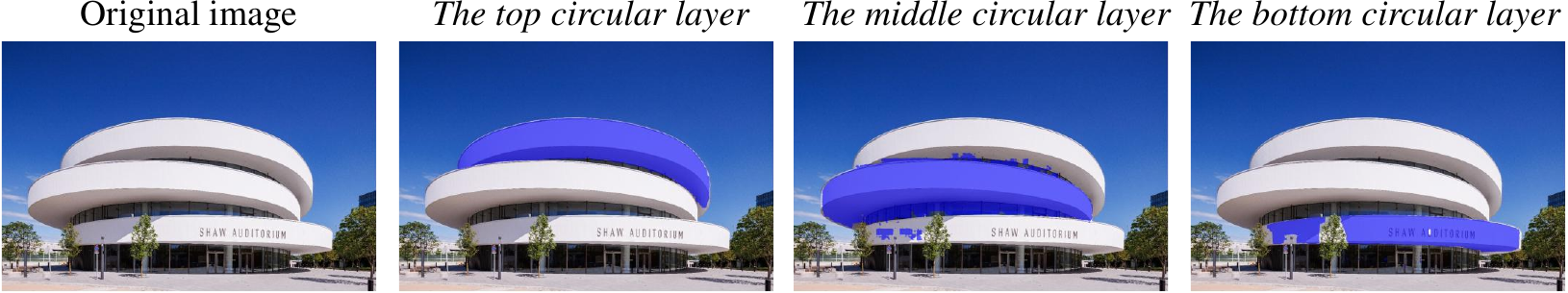}
    \caption{A complicated example of modern architecture.}
    \label{fig:shaw}
\end{figure*}
\begin{figure*}
    \centering
    \includegraphics[width=.9\linewidth]{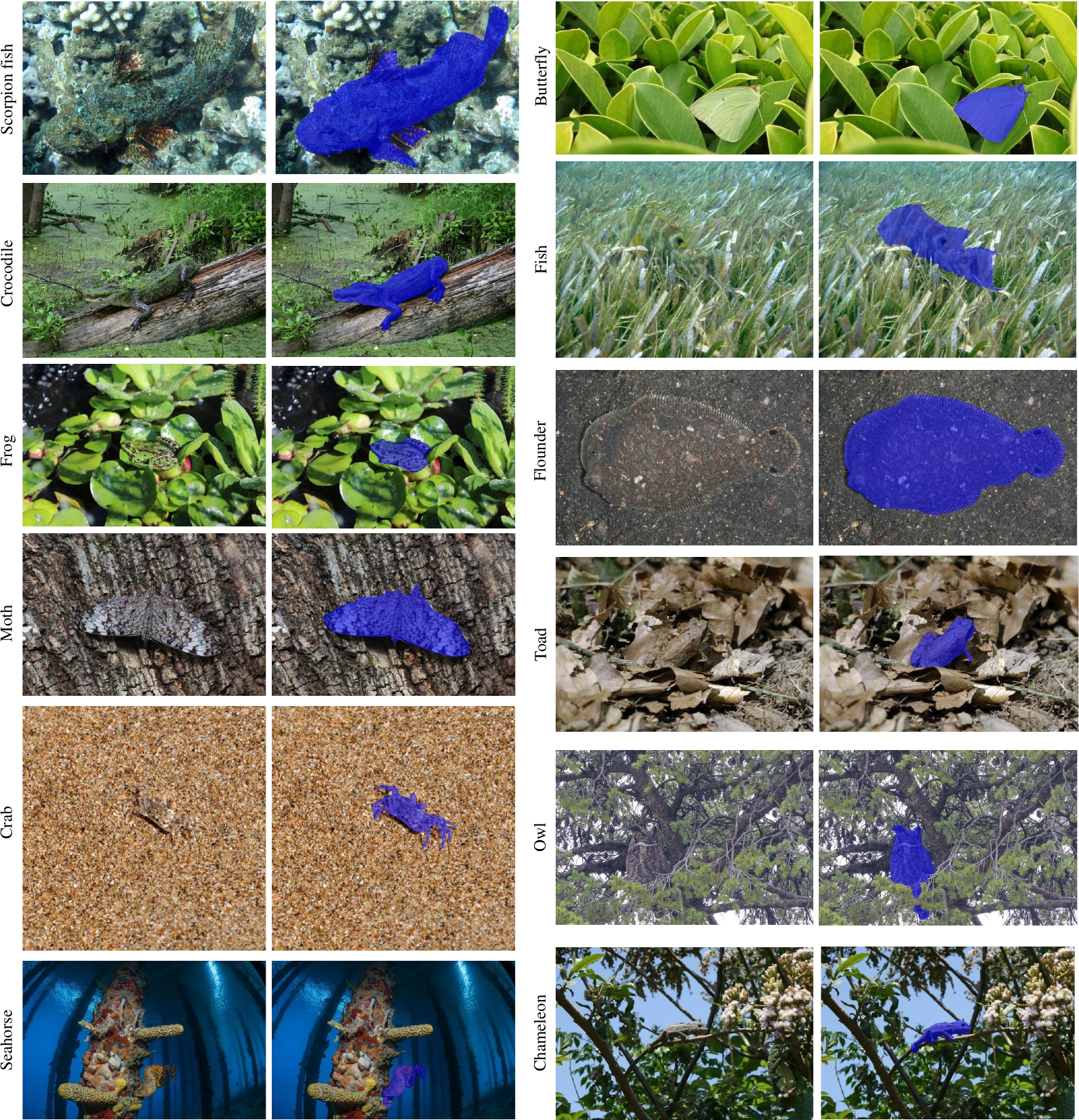}
    \caption{More camouflaged examples. Both explicit and implicit queries are used. Image classes are indicated next to the input images.}
    \label{fig:more}
\end{figure*}
\begin{figure*}
    \centering
    \includegraphics[width=.9\linewidth]{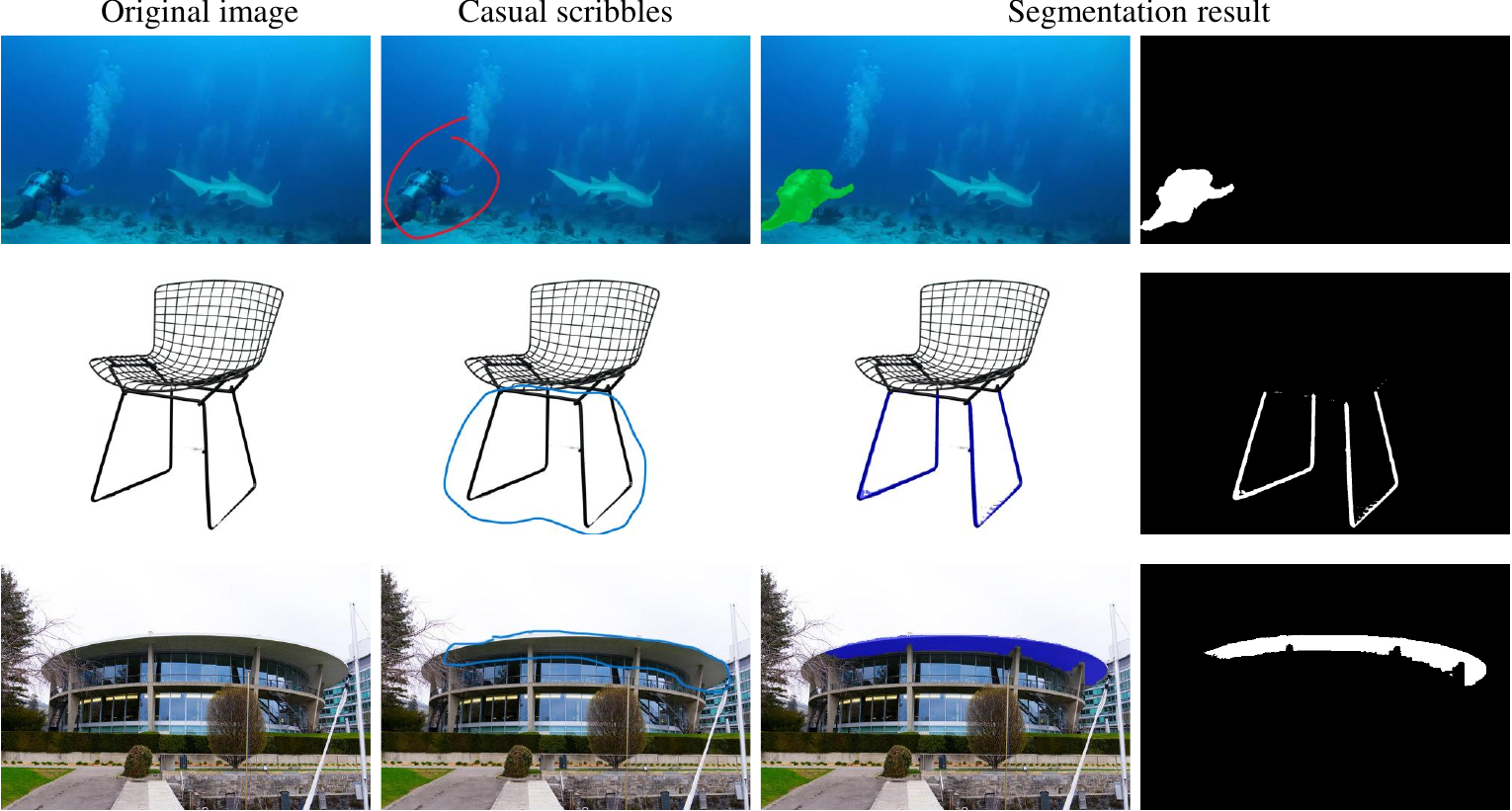}
    \caption{More results on scribble control.}
    \label{fig:more_scribble}
\end{figure*}

\paragraph{Casual scribbles.} We show more examples with casual scribble control in Figure~\ref{fig:more_scribble}. 

\begin{figure*}
    \centering
    \includegraphics[width=0.9\linewidth]{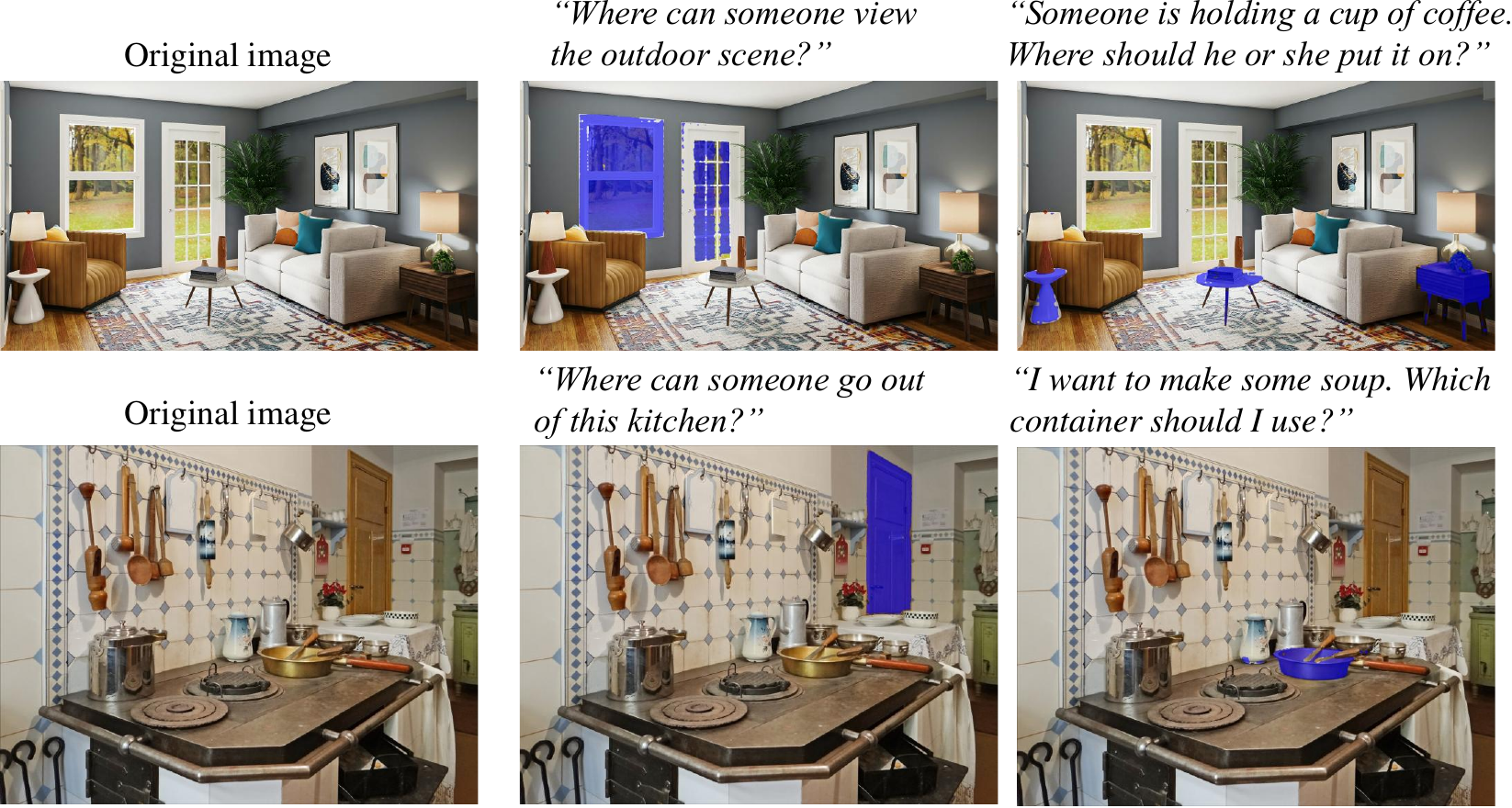}
    \caption{Results with more implicit queries.}
    \label{fig:implicit}
\end{figure*}

\paragraph{More implicit queries.} Figure~\ref{fig:implicit} demonstrates some examples with more implicit queries in real situations. Given the same scene, the user (robot) asks questions successively, from first viewing the scene (image) on the left, after Thinking First and then asking the first question in the middle, and asking the second question in the right of the figure. ThinkFirst in response produces the relevant segmentations in succession, which in practice sets the next targets for robotic navigation. 

\paragraph{Modern architecture.} We show a challenging example of modern architecture in Figure~\ref{fig:shaw}.

\begin{figure}
    \centering
    \includegraphics[width=\linewidth]{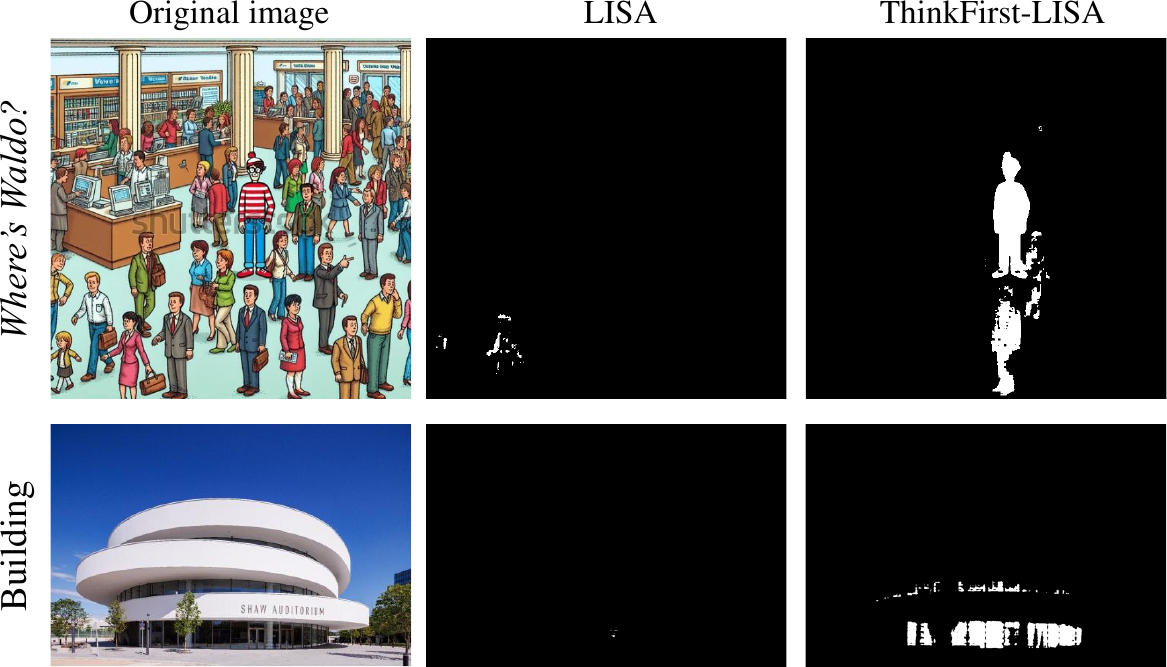}
    \caption{Failure cases of ThinkFirst. We compare the failure cases of ThinkFirst with LISA, while ThinkFirst can still be refined with casual scribbles. The first row includes an example of Where's Waldo, and the second row relates to a modern building with the prompt \textit{``Please segment the glass at the ground floor.''}.}
    \label{fig:fail}
\end{figure}

\paragraph{Failure cases.} Some failure cases of ThinkFirst are demonstrated in Figure~\ref{fig:fail}. They can be further refined with user's scribbles, as we provided-based control, using the pipeline in Figure 6.


\section{Concurrent Work}\label{app:concurrent}
We discuss the difference between our work and the concurrent work Seg-Zero~\cite{liu2025seg}. To our best knowledge, Seg-Zero introduces a robust and novel way to train multimodal LLM on outputing ``numbers''. In specific, if we simply ask GPT-4o to generate a bounding box or a point coordinate, the result is often inaccurate and discrete (e.g. with values like 100, 150, 200, etc). This is because existing MLLMs are good at text outputs instead of precision numbers. By using reinforcement learning (RL) with IoU rewards, Seg-Zero achieves impressive visual results on complex reasoning segmentation datasets. However, as shown in Figure~\ref{fig:segzero}, Seg-Zero fails to process simple cases for multiple objects of interest, due to the limitation of their RL dataset and rewarding mechanisms, thereby largely reducing the usability and applications of their framework. On the other hand, ThinkFirst demonstrates clearly better flexibilities, which are zero-shot without additional learning, much less  the notorious RL training process that possibly complicates the problem when CoT can be achieved at inference time with zero-shot capabilities.
{
    \small
    \bibliographystyle{ieeenat_fullname}
    \bibliography{main}
}

\end{document}